\documentclass[12pt, draftclsnofoot, onecolumn]{IEEEtran}
\IEEEoverridecommandlockouts
\usepackage{cite}
\usepackage{amsmath,amssymb,amsfonts}
\usepackage[linesnumbered,ruled,vlined]{algorithm2e}
\usepackage{graphicx}
\usepackage{epstopdf}
\usepackage{textcomp}
\usepackage{xcolor}
\usepackage{multirow}
\usepackage{soul}
\usepackage{array}

\usepackage[justification=centering]{caption}   

\usepackage{booktabs}

\usepackage{booktabs}
\def\BibTeX{{\rm B\kern-.05em{\sc i\kern-.025em b}\kern-.08em
    T\kern-.1667em\lower.7ex\hbox{E}\kern-.125emX}}
\begin{document}

\title{Mobility-Aware Asynchronous Federated Learning for Edge-Assisted Vehicular Networks}



\author{

{ 
	Siyuan Wang, Qiong Wu,~\IEEEmembership{Senior Member,~IEEE}, Qiang Fan,\\ Pingyi Fan, ~\IEEEmembership{Senior Member,~IEEE}, Jiangzhou Wang,~\IEEEmembership{Fellow,~IEEE}
}

\thanks{

{	
	
	Qiong Wu and Siyuan Wang are with the School of Internet of Things Engineering, Jiangnan University, Wuxi 214122, China, and also with the State Key Laboratory of Integrated Services Networks (Xidian University),  Xi'an 710071, China (e-mail: qiongwu@jiangnan.edu.cn, siyuanwang@stu.jiangnan.edu.cn)

	Qiang Fan is with Qualcomm, San Jose, CA 95110, USA (e-mail: qf9898@gmail.com)

	Pingyi Fan is with the Department of Electronic Engineering, Beijing National Research Center for Information Science and Technology, Tsinghua University, Beijing 100084, China (Email: fpy@tsinghua.edu.cn)

	Jiangzhou Wang is with the School of Engineering, University of Kent, CT2 7NT Canterbury, U.K. (Email: j.z.wang@kent.ac.uk)

}

}
}

\maketitle

\begin{abstract}
Vehicular networks enable vehicles support some real-time applications through training data. Due to the limited computing capability of vehicles, vehicles usually transmit data to a road side unit (RSU) deployed along the road to process data collaboratively. However, vehicles are usually reluctant to share data with each other due to the inevitable data privacy. For the traditional federated learning (FL), vehicles train the data locally to obtain a local model and then upload the local model to the RSU to update the global model through aggregation, thus the data privacy can be protected through sharing model instead of raw data. The traditional FL requires to update the global model synchronously, i.e., the RSU needs to wait for all vehicles to upload local models to update the global model. However, vehicles may usually drive out of the coverage of the marked RSU before they obtain their local models through training, which reduces the accuracy of the global model. In this paper, a mobility-aware vehicular asynchronous federated learning (AFL) is proposed to solve this problem, where the RSU updates the global model once it receives a local model from a vehicle where the mobility of vehicles, amount of data and computing capability are taken into account. Simulation experiments validate that our scheme outperforms the conventional AFL scheme.
\end{abstract}


\section{Introduction}
With the development of vehicular networks, vehicles can support the real-time applications such as speech recognition and multimedia sharing to facilitate people's life\cite{b1}.
In the vehicular networks, vehicles collect data from surroundings or networking and train the models based on raw data to support the vehicular services.
Due to the limited computing capability, vehicles usually transmit data to a fog or cloud with powerful computing, large storage and strong processing capability, then the fog or cloud processes the data and sends back the computing results to the corresponding vehicles in the vehicular networks. The fog or cloud is usually far from the vehicles, and thus incur a large delay due to the long transmitting distance.
To solve this issue, edge-assisted vehicular networks are introduced to process the data at a road side unit (RSU) deployed along the road in order to reduce the delay. However, vehicles are usually reluctant to share data with each other due to data privacy, thus the RSU may not be able to collect enough data to train an accurate model for vehicular services.

Federated learning (FL) provides a possible way to solve the above problem \cite{b2}. For the traditional FL, the RSU updates the global model for multiple rounds. For each round, each vehicle first downloads the global model from the marked RSU, then trains its local model based on the downloaded global model parameters. After that, each vehicle uploads its new updated local model to the RSU to further update the global model through aggregation. Thus the data privacy can be protected through sharing model parameters instead of raw data.

It is known that the traditional FL requires to update the global model synchronously, i.e., the RSU needs to wait for all vehicles to upload local models to update the global model in each round. However, vehicles may usually drive out of the coverage of the RSU before they get their own new update of local models. Therefore, the RSU cannot aggregate them of the vehicles to update the global model on time, resulting in reduction of the accuracy of the global model. The asynchronous federated learning (AFL) is essentially developed to solve this problem, where the RSU updates the global model once a vehicle uploads its local model, thus the global model can be updated in time without waiting for other vehicles to upload the local model.

In the edge-assisted vehicular networks, the positions of vehicles are time-varying due to the mobility of vehicles. In addition, the amount of data and computing capabilities of different vehicles may be different. These factors will affect the accuracy of the global model for the AFL. Specifically, the time varying position of vehicles will cause different transmitting rates and thus cause different uploading delays, i.e., the time to upload the local model. In addition, the vehicles with different computing capabilities and amounts of data will cause different local training delays, i.e., the time to train each local model. If the time duration of one vehicle from the global model downloading to the local model uploading is relatively large, the RSU may not ignore its role in the current epoch and update the global model based on the other vehicles. In this case, the local model of the vehicle is stale and not accurate, which further delays the convergence of FL and may reduce the accuracy of the global model.
Therefore, it is necessary to consider the mobility of vehicles, the amount of data and computing capability in designing an AFL scheme in the edge-assisted vehicular networks.

In this paper, we jointly consider the mobility of vehicles, amount of data and computing capability to design an mobility-aware AFL (MAFL) scheme\footnote{The source code has been released at: https://github.com/qiongwu86/AFLweight}. The contributions of this paper are summarized as follows.
{\begin{itemize}
\item[1)] We consider the mobility of vehicles, the different computing capabilities and amount of data of different vehicles to propose a mobility-aware AFL (MAFL) scheme to get a higher accuracy global model.
\item[2)] We compare our scheme with the traditional AFL to validate the outperformance of our scheme.
\item[2)] We investigate the performance of our new proposed scheme under the different aggregation proportion between the local model and global model.
\end{itemize}

The rest of this paper is organized as follows. Section II reviews the related work. Section III presents the system model. Section IV presents our scheme in detail. Section V shows the simulation results. Section VI concludes this paper.

\section{Related Work}
In recent years, some works have studied FL in edge-assisted vehicular networks.
In \cite{b3}, Xiao \emph{et al.} considered the mobility of vehicles to jointly optimize the onboard computing capability, transmission power and local model accuracy to minimize the cost of the FL.
In \cite{b4}, Yuan \emph{et al.} proposed a bidirectional connection broad learning system to train the dataset stored in vehicles and then proposed a federated broad learning system algorithm to improve the aggregation efficiency.
In \cite{b5}, Zhou \emph{et al.} employed a FL in a two-layer vehicular network and designed a novel multi-layer heterogeneous model selection and aggregation scheme to achieve high accuracy. Moreover, they designed a context-aware learning mechanism to reduce the learning time of the object detection.
In \cite{a1}, He \emph{et al.} proposed a scheme that considered both FL and blockchain to protect the privacy of vehicles and prevent the possible malicious attacks.
In \cite{a2}, Ayaz \emph{et al.} employed FL and blockchain among vehicles to improve the message delivery rate while protecting the privacy of vehicles.
In \cite{a3}, Zhang \emph{et al.} proposed a method using federated transfer learning to detect the drowsiness of drivers while protecting their privacy.
In \cite{a4}, Yu \emph{et al.} proposed a proactive edge caching scheme using FL while considering the mobility of vehicles to improve the cache hit ratio.
In \cite{a5}, Wang \emph{et al.} proposed a scheme using deep learning to collect and preprocess the data for FL. The scheme can reduce the amount of the data that upload to the cloud and protect the privacy of users effectively.
In \cite{a6}, Yan \emph{et al.} proposed a distributed power allocation scheme to maximize the energy efficiency and protect the privacy of users using FL.
In \cite{a7}, Chai \emph{et al.} proposed a hierarchical algorithm to improve the efficiency of knowledge sharing and can prevent some malicious attack using FL and blockchain.
In \cite{a8}, Zhao \emph{et al.} designed a protocol for data sharing to get the lower packet loss rate and authentication delay by using FL.
In \cite{a9}, Ye \emph{et al.} proposed a selective model aggregation approach to get a higher accuracy of global model.
In \cite{a10}, Zhao \emph{et al.} proposed a scheme combined FL with local differential privacy.
However, these works adopt the traditional FL, which needs the RSU to wait all vehicles to upload their models and then update the global model. Vehicles may usually drive out of the coverage of the RSU before they upload their local models, which deteriorates the accuracy of the global model when the traditional FL is adopted.

A few works have studied the AFL in vehicular networks.
In \cite{b07}, Tian \emph{et al.} proposed an asynchronous federated deep Q-learning network to solve the task offloading problem in vehicular network, then designed a queue-aware algorithm to allocate computing resources.
In \cite{b09}, Pan \emph{et al.} proposed a scheme using AFL and deep Q-learning algorithm to get the maximized throughput while considering the long-term ultra-reliable and low-latency communication constraints.
However, these works have not considered the mobility of vehicles, the amount of data and computing capability in the design of the AFL in vehicular networks.

As mentioned above, there is no work jointly considering the mobility of vehicles, amount of data and computing capability to design the AFL to improve the accuracy of global model in vehicular networks.

\section{System Model}
In this section, we first describe the scenario, and then briefly review the process of AFL.

\subsection{scenario}\label{AA}
As shown in Fig. \ref{fig1}, we consider an edge-assisted vehicular network architecture consisting of one RSU at the network edge and $K$ vehicles driving with a constant velocity in the coverage of the marked RSU. All vehicles are driving to the east. Different vehicles collect different amounts of data and their computing capabilities may be also different. The traversing time duration within the RSU's coverage is divided into discrete time slot. A three dimensional coordinate system is constructed to facilitate the formulation of vehicle position in real time, where the origin is set as the position of the bottom of the RSU, the direction of $x$ axis is set as east and the direction of $y$ axis is south, the direction of $z$ axis is set along the antennas of RSU which are perpendicular to both $x$ and $y$ axis.

\begin{figure}
\center
\includegraphics[scale=0.55]{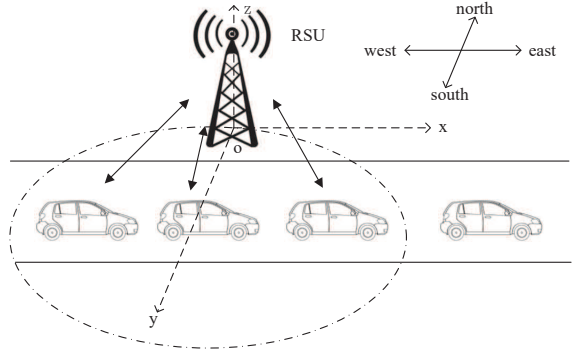}
\caption{System model}
\label{fig1}
\end{figure}

\subsection{process of AFL}\label{AA}
The vehicular network adopts the AFL to update model for $M$ rounds. For the first round, the RSU initializes the global model based on the convolutional neural network (CNN) and each vehicle downloads the global model from the RSU. Then each vehicle adopts the global model to update the local model through training its local data based on CNN and uploads its local model after finishing its local training.  The RSU updates the global model immediately once it receives a local model from a vehicle. At this moment, the first round training is finished. After that the vehicle downloads the global model from the RSU to continue to update its local model. The RSU would update the global model again when it receives another local model. The above procedures will be repeated for many rounds. When the global model is updated for $M$ rounds, the AFL is finished and the RSU obtains the final global model.

\begin{figure}
\center
\includegraphics[scale=0.55]{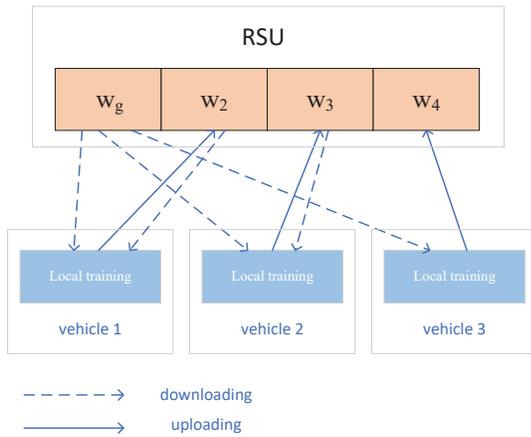}
\caption{The process of AFL}
\label{sys2}
\end{figure}

The process of AFL is shown in Fig. \ref{sys2}, the RSU first initializes the global model as $w_{g}$, then each vehicle downloads $w_{g}$ for local training. After that, each vehicle sends its local model to the RSU. We assume that vehicle 1 first finishes the local training, vehicles 2 and vehicle 3 are finished later. Thus RSU first receives the local model of vehicle 1 and then updates the global model as $w_{2}$, then vehicle 1 downloads $w_{2}$ from the RSU for the next local training.
After that, RSU receives the local model of vehicle 2 and vehicle 3, respectively, and updates the global model in the same way.

\section{MAFL Scheme for Edge-Assisted Vehicular Networks}
In this section, we propose a MAFL scheme to improve the accuracy of the global model in edge-assisted vehicular networks. The scheme is executed for $M$ rounds. We consider in round $r$ $({r \in \left[ {1,M} \right]})$, vehicle $i$ experiences the process including global model downloading, local training and uploading the local model. Next we will introduce the process of round $r$ for vehicle $i$ in detail.

\subsection{Downloading Global Model}\label{AA}
For round $r$, vehicle $i$ first downloads the global model $w_{r-1}$ from the RSU after it updates the global model based on the local model of vehicle $i$ in round $r-1$. Particularly, at the first round of the scheme, the global model $w_{g}$ is initialized by the RSU based on the CNN .

\subsection{Local Training}\label{AA}
Then vehicle $i$ employs the global model $w_{r-1}$ to update its local model through training. Note that $w_{r-1}$ is $w_{g}$ in the first round. Specifically, the training process consists of $l$ iterations. Let $w_{r,j}^{i}$ be the local model parameter for vehicle $i$ in iteration $j$ $({j \in \left[ {1,l} \right]})$ of round $r-1$, which is updated at the end of iteration $j-1$. For each iteration $j$, vehicle $i$ first inputs the probability of the label of each local data $a$, denoted as ${{y}_{a}}$, into the CNN with local model $w_{r,j}^{i}$ and outputs the predicted probability of the label of each data $a$, denoted as $\hat{{y}_{a}}$.

Then vehicle $i$ adopts the cross-entropy loss function to calculate the loss of $w_{r,j}^{i}$, i.e.,

\begin{equation}
{{f}_{i}}\left( w_{r,j}^{i} \right)\text{=}-\sum\limits_{a=1}^{{{D}_{i}}}{{{y}_{a}}\log }\hat{{y}_{a}}\,,
\label{eq1}
\end{equation}
where ${{D}_{i}}$ is the number of data carried by vehicle $i$.

Then the stochastic gradient descent algorithm is adopted to update the local model, i.e.,
\begin{equation}
w_{r,j+1}^{i}=w_{r,j}^{i}-\eta \nabla {{f}_{i}}\left( w_{r,j}^{i} \right),
\label{eq2}
\end{equation}
where $\nabla {{f}_{i}}\left( w_{r,j}^{i} \right)$ is the gradient of ${{f}_{i}}\left( w_{r,j}^{i} \right)$ and $\eta $ is the learning rate. Then vehicle $i$ sets $w_{r,j+1}^{i}$ as the model of the CNN and starts iteration $j+1$ to update the local model. When the number of iterations reaches $l$, vehicle $i$ obtains the updated local model $w_{r}^{i}$.


Different from the AFL, we consider the mobility of vehicles, amount of data and computing capability to calculate the weighted local model $w_{r_u}^{i}$. Specifically, on one hand, the mobility of vehicles is considered to calculate the time varying transmission distance, then the time varying transmission rate is calculated based on the transmission distance, afterwards, the time varying uploading delay and uploading delay weight are further obtained based on the transmission rate. On the other hand, the amount of data and computing capability are considered to calculate the local training delay, then the local training delay weight is obtained based on the local training delay. After the uploading delay weight and local training delay weight are calculated, the weighted local model $w_{r_u}^{i}$ is further calculated based on the two weights. The process to calculate the weighted local model is introduced as follows.

The uploading delay weight is first calculated. Let ${{P}^{i}}\left( t \right)$ be the position of vehicle $i$ at time slot $t$ and it is expressed as $\left( d_{x}^{i}\left( t \right),d_{y}^i,0 \right)$,  $d_{x}^{i}\left( t \right)$ and ${d_{y}^{i}}$ are the distances between vehicle $i$ and the antennas of RSU along $x$ axis and $y$ axis at slot $t$, respectively, and ${d_{y}^{i}}$ is a fixed $d_{y}$ for each vehicle $i$. Since each vehicle drives at a constant speed ${v}$, $d_{x}^{i}(t)$ is calculated as
\begin{equation}
{d_{x}^{i}\left( t \right)}=d_{x}^{i}(0)+{{v}}\cdot t,
\label{eq12}
\end{equation}
where $d_{x}^{i}(0)$ is the position of vehicle $i$ at the initial time slot, which is known by vehicle $i$ based on the global positioning system (GPS).

Let ${{d}^{i}}\left( t \right)$ be the distance between vehicle $i$ and the RSU at time slot $t$. The height of the RSU is $H$, thus the position of the RSU's antennas ${{P}_{R}}$ is $\left( 0,0,H \right)$. In this case, ${{d}^{i}}\left( t \right)$ is calculated as
\begin{equation}
{{d}^{i}}\left( t \right)=\left\| {{P}^{i}}\left( t \right)-{{P}_{R}} \right\|,
\label{eq13}
\end{equation}

According to Shannon's theorem, the transmission rate of vehicle $i$ at time slot $t$ is calculated as
\begin{equation}
r_{r,u}^i\left( t \right)=B{{\log }_{2}}\left( 1+\frac{p_{m}\text{ }\!\!\cdot\!\!\text{ }{{h}^{i}}\text{ }\!\!\cdot\!\!\text{ }{{\left( {{d}^{i}}\left( t \right) \right)}^{-\alpha }}}{{{\sigma }^{2}}} \right),
\label{eq8}
\end{equation}
where $B$ is the bandwidth, $p_{m}$ is the transmit power of each vehicle, ${{h}^{i}}$ is the channel gain of vehicle $i$, $\alpha$ is the path loss exponent and ${{\sigma }^{2}}$ is the noise power.

Let $C_{r,u}^{i}(t)$ be the uploading delay of vehicle $i$ at time slot $t$ and thus it is calculated as
\begin{equation}
C_{r,u}^{i}\left( t \right)=\frac{\left| {{w}} \right|}{r_{r,u}^i\left( t \right)},
\label{eq7}
\end{equation}
where $\left| {{w}} \right|$ is the size of the local model of each vehicle, and ${r_{r,u}^i}\left( t \right)$ is the transmission rate of vehicle $i$ at time slot $t$.

If the uploading delay is large, the local model would be stale and thus not accurate, in this case the uploading delay weight $\beta _{r,u}^{i}\left( t \right)$ should be small. Hence, the uploading delay weight at time slot $t$ is calculated as

\begin{equation}
\beta _{r,u}^{i}\left( t \right)={{\gamma }^{C_{r,u}^{i}\left( t \right)-1}},
\label{eq6}
\end{equation}
where $\gamma$ is a parameter which belongs to $\left( 0,1 \right)$ to make $\beta _{r,u}^{i}\left( t \right)$ decrease as the uploading delay increases.

Then the local training delay weight is calculated. Let $C_{l}^{i}$ be the local training delay of vehicle $i$, thus it is calculated as
\begin{equation}
C_{l}^{i}=\frac{{{D}_{i}}\text{ }\!\! \cdot \! \! C_y}{{{\delta }_{i}}},
\label{eq5}
\end{equation}
where $C_y$ is the required CPU cycle to train a data and ${{\delta }_{i}}$ is the CPU cycle frequency allocated for vehicle $i$.

Similar with the uploading delay weight, the local training delay weight $\beta _{r,l}^{i}$ is calculated as

\begin{equation}
\beta _{r,l}^{i}={{\zeta }^{C_{l}^{i}-1}},
\label{eq4}
\end{equation}
where $\zeta$ is a parameter which belongs to $\left( 0,1 \right)$ to make $\beta _{r,l}^{i}$ decrease as the local training delay increases.

After the uploading delay weight and local training delay weight are obtained. The weighted local model $w_{r_u}^{{{i}}}$ is calculated as
\begin{equation}
w_{r_u}^{{{i}}}=w_r^{i}*\beta _{r,u}^{i}(t)*\beta _{r,l}^{i},
\label{eq9}
\end{equation}
Note that the download bandwidth is large, thus the downloading delay is much small and can be neglected here.



%

\subsection{Uploading Local Model and Aggregation}
After that vehicle $i$ uploads the weighted local model $w_{r_u}^{{{i}}}$ to the RSU, then the RSU updates the global model through aggregation. The global model in round $r$ is updated as

\begin{equation}
{{w}_{r}}=\beta {{w}_{r-1}}+\left( 1-\beta  \right)w_{r_u}^{{{i}}},
\label{eq10}
\end{equation}
where ${{w}_{r-1}}$ is the global model updated at the end of round $r-1$, $\beta $ is the aggregation proportion that belongs to $\left( 0,1 \right)$.

Up to now, round $r$ is finished. Then the scheme moves to round $r+1$ to update the global model. When the number of rounds reaches $M$, the RSU finishes updating its global model and achieves the final global model. The detailed process of the MAFL scheme is shown in Algorithm 1.

\begin{algorithm}
	\small
	\caption{Mobility-awared asynchronous federated learning scheme}
	\label{al1}
	Initialize the global model $w_g$;\\
	\For{each round $r$ from $1$ to $M$}
	{
		$w_r^i \leftarrow \textbf{Vehicle Updates}(w_g)$;\\
		Vehicle $i$ calculates the weighted local model $w_{r_u}^i$ based on Eq. \eqref{eq9};\\
		Vehicle $i$ uploads the weighted local model $w_{r_u}^i$ to the RSU;\\

		RSU receives the weighted local model $w_{r_u}^i$ from vehicle $i$;\\
		RSU updates the global model based on Eq. \eqref{eq10};\\
		\Return $w_r$
	}
	\textbf{Vehicle Update}($w$):\\
	\textbf{Input:} $w_g$ \\
	\For{each local iteration $j$ from $1$ to $l$}
	{
		Vehicle $i$ calculates the cross-entropy loss function based on Eq. \eqref{eq1};\\
		Vehicle $i$ updates the local model $w_{r,j}^i$ based on Eq. \eqref{eq2};\\
	}
	Set $w_r^i=w_{r,j}^i$;\\
	\Return$w_r^i$
	
\end{algorithm}

\section{Simulation Results}
In this section, we conduct simulation experiments to verify the effectiveness of the proposed scheme. We will first introduce the relevant simulation settings and evaluate the performance of our scheme.

\subsection{Simulation Setup}

\begin{table}\footnotesize
\caption{Values of parameters}
\label{tab1}
\centering
\begin{tabular}{|c|c|c|c|}
\hline
\textbf{Parameter} &\textbf{Value} &\textbf{Parameter} &\textbf{Value}\\
\hline
$K$ & 10 & $v$ & $20m/s$ \\
\hline
$H$ & $10m$ & $d_{y}$ & $10m$ \\
\hline
$C_y$ & ${{10}^{5}}cycles$ & $\left| {{w}} \right|$ & $5000bits$ \\
\hline
$B$ & ${{10}^{5}}HZ$ & $p_{m}$ & $0.1w$ \\
\hline
$\alpha$ & 2 & ${{\sigma }^{2}}$ & ${{10}^{-11}}mw$ \\
\hline
$\beta$ & 0.5 & $\zeta$ & 0.9 \\
\hline
$\gamma$ & $0.9$ & $ $ &  \\
\hline
\end{tabular}
\end{table}

The simulation tool is Python 3.9. ${\delta_{i}}$ is set as $1.5\cdot \left( i+ 5 \right)\cdot {{10}^{8}}$ cycles/s. We employ the orthogonal frequency division multiplexing (OFDM) technology and Rayleigh fading channel in this simulation \cite{wj1}, \cite{wj2}, thus $h_i$ is a stochastic value for each vehicle $i$ based on the autoregressive model \cite{b9}. The data carried by each vehicle are the images randomly selected from 60000 images in the MNIST dataset, where vehicle $i$ carries $\left( 2250+3750i \right)$ images. The detailed simulation parameters are listed in Table \ref{tab1}.

\subsection{Performance Evaluation}
The accuracy (i.e., denoted as acc) and loss of the global model are adopted as the metrics in our experiments, where the loss is calculated according to Eq. \eqref{eq1} and the accuracy reflects the classification performance of the global model obtained by our scheme and is defined as
\begin{equation}
acc=\frac{\text{the number of correctly classified data}}{\text{the number of total data}}\times 100\%
.
\label{eq11}
\end{equation}

Next, we evaluate the metrics of the global model in our scheme by comparing with the traditional AFL scheme, where each result is obtained by averaging the results of three experiments.
\begin{figure}
\center
\includegraphics[scale=0.55]{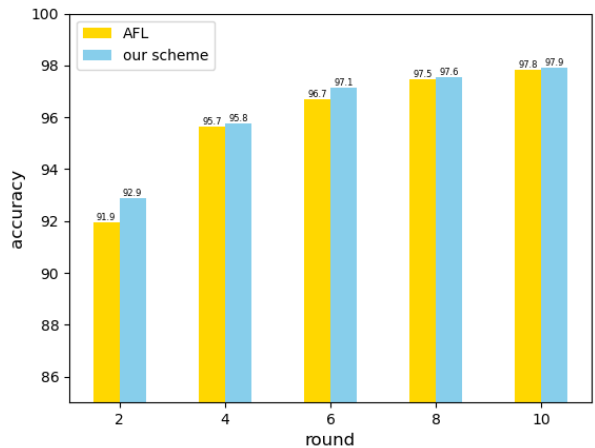}
\caption{Accuracy comparison of AFL and our proposed scheme in different rounds}
\label{fig2}
\end{figure}

\begin{figure}
\center
\includegraphics[scale=0.55]{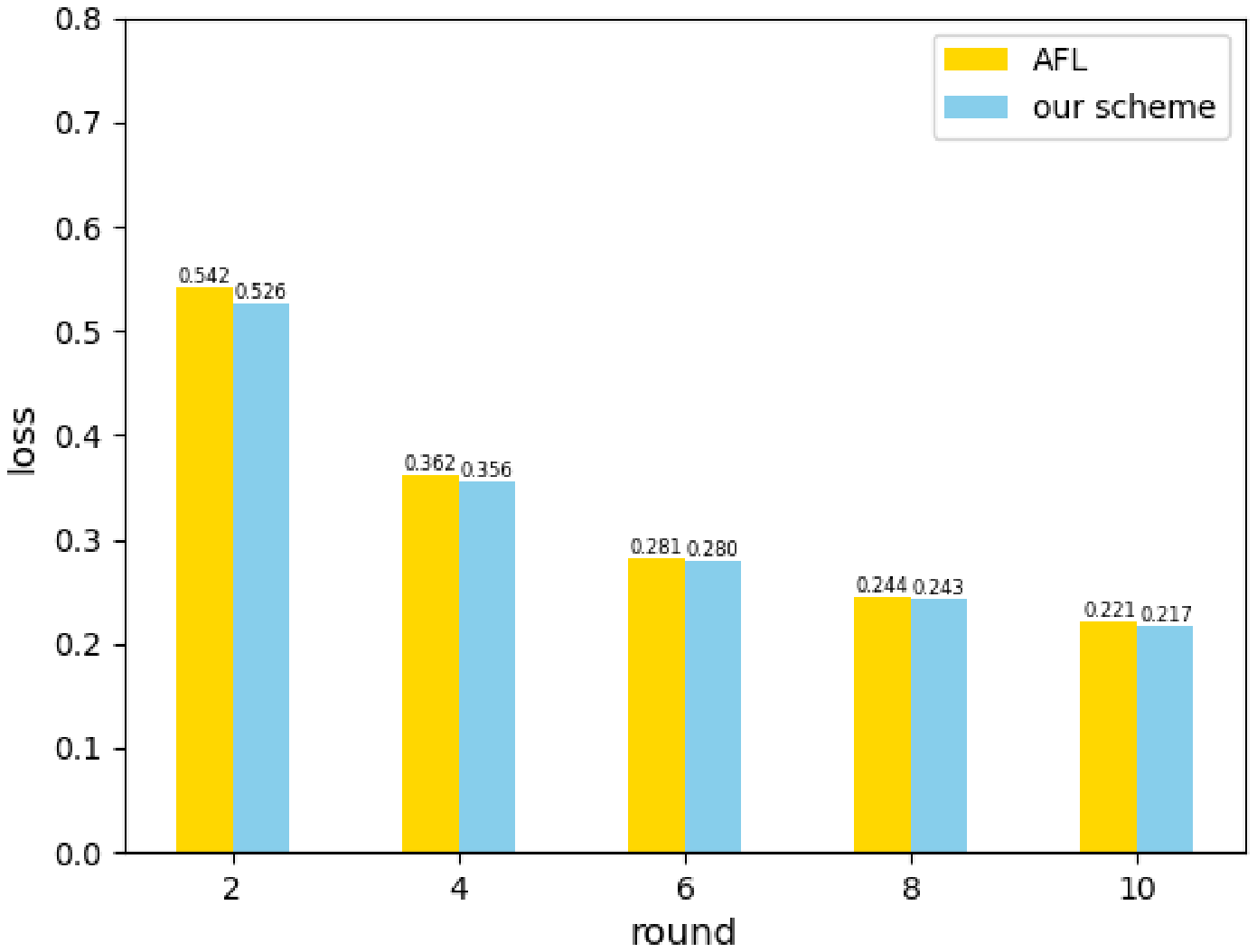}
\caption{Loss comparison of AFL and our proposed scheme in different rounds}
\label{fig3}
\end{figure}

\begin{figure}
\center
\includegraphics[scale=0.55]{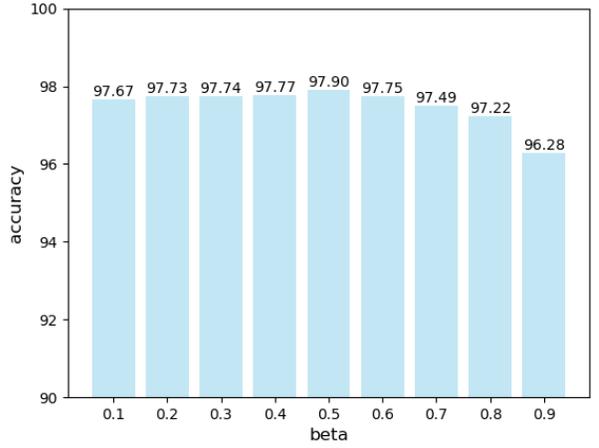}
\caption{Accuracy under different beta}
\label{fig4}
\end{figure}

Fig. \ref{fig2} depicts the accuracy of the global models under the AFL and our scheme in different rounds. We can observe that the accuracy under both schemes increases as the number of rounds increases, and finally becomes relatively stable.
This is because the global model is updated by more local models with the number of rounds increases, thus the accuracy of the global model is improved. When the number of rounds continues to increase, the RSU has updated the global model through aggregating enough local models to obtain the accurate global model, thus the accuracy of the global model becomes stable.
In addition, our scheme can finally achieve relatively higher accuracy. This is because our scheme considers amount of data, computing capability and mobility of vehicles to improve the accuracy of the global model.

Fig. \ref{fig3} depicts the loss of the global models under AFL and our scheme in different rounds. We can see that the loss under the two schemes decreases as the number of rounds increases. This is because the accuracy under the two schemes increases as the number of rounds increases. Moreover, we can see that our scheme has a lower loss due to that our scheme synthetically considers multiple factors to improve the accuracy of the global model.

Fig. \ref{fig4} shows the model accuracy of our scheme under different $\beta$ when the number of rounds is $10$. Our scheme keeps a relatively high accuracy when $\beta$ is smaller than 0.5. In addition, the accuracy decreases as the value of $\beta$ increases when $\beta$ is larger than 0.5. This is because according to Eq. \eqref{eq10}, the weight of the local model is small when $\beta$ is larger than 0.5. In this case, the global model tends to be updated according to the previous global model, which decreases the accuracy. When $\beta$ is 0.9, the accuracy is decreased significantly. This is because the weight of the local model is much small, thus the update of the global model majorly depends on the previous global model, which decreases the influence and contribution of new data of all the vehicles.


\section{Conclusions}
In this paper, we have considered the amount of data, computing capability and vehicle mobility and proposed a MAFL scheme to improve the accuracy of global model. We have conducted extensive simulations to demonstrate the performance of the proposed scheme. The conclusions are summarized as follows:
\begin{itemize}
    \item Our scheme considered the amount of data, computing capability and mobility of vehicles to improve the accuracy and thus outperforms the AFL scheme.	

    \item With the increase of training rounds, RSU receives more local model from vehicles to update global model, thus the accuracy of global model is increasing and the loss of global model is decreasing. Finally, the training becomes convergent and thus both the accuracy and loss of the global model becomes stable.

	\item The aggregate proportion $\beta$ affects the accuracy of the global model and the desirable accuracy is achieved when $\beta$ is relatively small.
\end{itemize}

\end{document}